%% file: main.tex
\documentclass[letterpaper]{article} 
\usepackage[preprint]{aaai2027}  
\usepackage[hyphens]{url}  
\usepackage{graphicx} 
\usepackage{natbib}  
\usepackage{caption} 
\usepackage{algorithm}
\usepackage{algorithmic}

\usepackage{newfloat}
\usepackage{listings}
\DeclareCaptionStyle{ruled}{labelfont=normalfont,labelsep=colon,strut=off} 
\floatstyle{ruled}
\newfloat{listing}{tb}{lst}{}
\floatname{listing}{Listing}

\usepackage{booktabs}

\usepackage{amsmath}
\usepackage{amssymb}
\usepackage{multirow}
\usepackage{makecell}
\usepackage{colortbl}
\usepackage{pifont}

\title{Parasitic Co-Denoising: Unlocking 3D Human Motion Generation in a Frozen Video Diffusion Model}
\author{
    Yunjiao Zhou,
    Junlang Qian,
    Lihua Xie,
    Jianfei Yang
}
\affiliations{
    Nanyang Technological University, Singapore
}

\begin{document}

\maketitle

\begin{abstract}
Despite never being supervised on explicit 3D motion, large-scale 
text-to-video diffusion models synthesize realistic human motion 
in their generated videos. We ask whether this implicit knowledge 
can be turned into explicit 3D motion generation, without training 
a separate motion model. Probing a frozen Wan2.1 reveals that a 
recoverable motion signal is present in its intermediate states 
across the entire denoising schedule, not confined to the clean 
output. Motivated by this, we introduce parasitic co-denoising, a paradigm in which motion is decoded from the host model along its denoising schedule rather than produced by an independent generator. We instantiate it as the Parasitic Motion Decoder (PMD), an efficient flow-matching decoder that shares the host's noise 
schedule and reads its intermediate features through a 
$\sigma$-adaptive multi-layer fusion, leaving the host unmodified. 
Drawing its coverage from the host rather than from motion data, PMD leads dedicated motion generators on 
text-motion alignment at a small fraction of their trainable 
parameters, while producing paired video and motion in a single 
pass that motion-only baselines cannot match.

\end{abstract}



\input{chapter/1intro}
\input{chapter/2relatedwork}
\input{chapter/3method}
\input{chapter/4experiment}
\input{chapter/5conclusion}

\bibliography{aaai2027}


\end{document}

%% file: chapter/1intro.tex
\section{Introduction}

Human motion generation~\cite{plappert2016kit, yu2026cmdm} drives applications such as animation~\cite{mourot2022survey}, virtual reality~\cite{burdea2003virtual}, and embodied AI~\cite{duan2022survey}. Recent text-to-motion models advance rapidly~\cite{weng2026realign, zeng2025lightt2m}, yet the scarcity of motion-capture data leaves them unable to cover the long tail of human behavior. Large-scale video diffusion models~\cite{blattmann2023stable}, by contrast, train on internet-scale data and render diverse humans in natural, varied motion. Such motion, learned without any explicit 3D supervision, shows that these models already internalize how humans move. This paper asks whether this implicit knowledge can become explicit 3D motion generation, without training a separate motion model.

Recent efforts leverage video models for 3D motion generation~\cite{yang2025echomotion, lin2025quest}, jointly training a separate motion branch under the guidance of video features. For this branch to inherit the video model's generalization, it must be trained on 3D motion at a comparable scale, which drives these methods to contribute large paired video-motion datasets. This effort rebuilds in a separate pipeline a capacity the video model may already possess, and incurs the additional cost of training and storing a second generative model. This suggests a more direct alternative: if motion is already encoded in the video model's generative process, can it be decoded from it, rather than relearned in a separate model?

\begin{figure*}[t]
  \centering
  \includegraphics[width=0.99\linewidth]{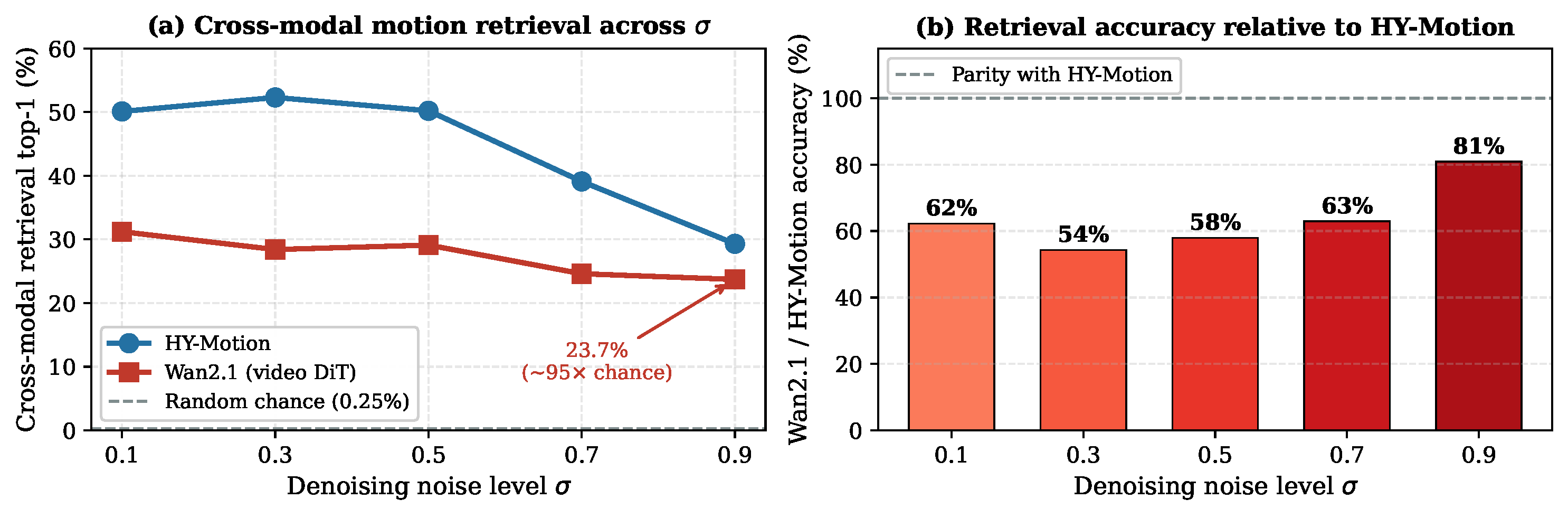}
  \vspace{-3mm}
  \caption{%
    Motion is decodable from a frozen video diffusion 
    model at every step of its denoising trajectory.
    Cross-modal retrieval top-1 accuracy between the Wan2.1 
    video velocity and paired 3D motion at each noise level 
    $\sigma$, against a held-out pool of 400 candidates 
    (chance $=0.25\%$), compared with HY-Motion, a large-scale 
    flow-matching model trained explicitly for 3D motion.
    \textbf{(a)} Frozen Wan2.1 retrieves motion at 24--31\% 
    across the schedule, two orders of magnitude above chance.
    \textbf{(b)} At every $\sigma$, Wan2.1 reaches 54--81\% of 
    HY-Motion's accuracy, peaking at $81\%$ at $\sigma=0.9$ 
    where the input is nearly pure Gaussian noise. 
  }
  \vspace{-5mm}
  \label{fig:probing}
\end{figure*}

To answer this, we first ask where motion should be decoded from. Reading from the model's clean output trivially recovers motion but reduces the problem to pose estimation~\cite{zhou2023metafi++}, ignoring the generative process that produced it. The question we pursue is whether motion is already present in the intermediate denoising states themselves. At each noise level $\sigma$, we probe a frozen Wan2.1 video model~\cite{wan2025wan} for the alignment between its intermediate state and the corresponding state of HY-Motion~\cite{hymotion2025}, a state-of-the-art open-source text-to-motion flow-matching model (Figure~\ref{fig:probing}). The two are remarkably aligned. Across the entire schedule, the video model's intermediate states retain 54--81\% of the dedicated model's motion-discriminative accuracy, holding 81\% even at $\sigma=0.9$, where the input is nearly pure noise. Motion information inside the video model is therefore not localized at its clean endpoint. It is present throughout the denoising trajectory, exposing a signal that a motion generator can ride on rather than relearn.

The signal exposed above suggests a separate motion generator is unnecessary, and motion can instead be decoded from the video model along its denoising schedule, produced together with the video. We call this strategy \emph{parasitic co-denoising}. By reusing the host's features at each step, the motion decoder attains the video model's broad coverage at little extra cost. This requires more than reading off features at a single point. Such a readout accesses only one slice of the host's trajectory and discards the rest, contradicting the very finding that motion is present at every noise level. To exploit that presence, the motion generator must itself unfold along the same trajectory, sharing the host's noise schedule and advancing one step at a time alongside it. The motion decoder therefore advances on the same schedule as the host and conditions each step on the host's matched-$\sigma$ features, so that video generation and motion decoding proceed in step.

We realize this design as the \textbf{Parasitic Motion Decoder} (PMD), a parameter-efficient decoder attached to a frozen Wan2.1 video DiT. PMD is itself a flow-matching decoder sharing the host's noise schedule. At each step, it reads intermediate features from the video DiT, fuses them through a $\sigma$-adaptive mechanism that learns which layers to attend to at each noise level, and updates its own motion latents via a learned velocity field. The video model is never modified. On open-world motion synthesis, PMD leads dedicated motion generators trained on substantially larger motion corpora, and inherits the broad semantic coverage of the underlying video model, generalizing to complex actions that lie outside the reach of motion-specific datasets.

The main contributions are summarized as follows:
\begin{itemize}
  \item We empirically show that a frozen text-to-video diffusion model already carries a recoverable 3D human motion signal, decodable from its intermediate states throughout the entire denoising schedule.
  
  \item We introduce \emph{parasitic co-denoising}, a paradigm in which a secondary modality is decoded from a host model along its own denoising schedule rather than produced by an independent generator, applicable to hosts whose denoising process exposes a usable signal for that modality.
  
  \item PMD leads dedicated motion generators on text-motion alignment while producing paired video and motion in a single pass, at a small fraction of trainable parameters.
\end{itemize}

%% file: chapter/2relatedwork.tex
\section{Related Work}

\paragraph{Text-to-motion generation.}
Methods for text-conditioned 3D human motion generation can be 
broadly grouped by representation. Early approaches encode motion 
in a variational latent~\cite{petrovich2022temos, guo2022tm2t}. 
Diffusion-based methods including MDM~\cite{tevet2022human} and 
MotionDiffuse~\cite{zhang2024motiondiffuse} denoise raw kinematic 
sequences, with later work adding physical 
plausibility~\cite{yuan2023physdiff} and fine-grained 
control~\cite{xie2024omnicontrol}. Token-based methods such as 
T2M-GPT~\cite{zhang2023generating} and MoMask~\cite{guo2024momask} 
learn discrete motion tokens with autoregressive or masked 
transformers, a recipe later scaled with language-model 
backbones~\cite{jiang2023motiongpt}. 
FineMoGen~\cite{zhang2023finemogen} introduces fine-grained 
spatio-temporal control, MotionLCM~\cite{dai2024motionlcm} 
accelerates inference via consistency distillation, and 
MotionCraft~\cite{bian2025motioncraft} scales motion training 
data. 

\paragraph{Reusing pretrained generative models.}
A growing line of work reuses features or weights of large-scale 
pretrained generative models for downstream tasks~\cite{yang2024cogvideox,kong2024hunyuanvideo}. Diffusion 
features have been shown to support semantic 
segmentation~\cite{baranchuk2021label}, semantic 
correspondence~\cite{tang2023emergent}, monocular 
depth~\cite{ke2024marigold}, self-supervised representation 
learning~\cite{xiang2023denoising}, and 
self-conditioning~\cite{feng2024latent}. In the motion domain, 
ViMoGen~\cite{lin2025quest} uses video features as auxiliary 
supervision when training a $1.2$B text-to-motion DiT (with a 
distilled $600$M variant for video-free inference); 
EchoMotion~\cite{yang2025echomotion} jointly fine-tunes video and 
motion branches of a dual-modality DiT. 
MoAlign~\cite{bhowmik2025moalign} aligns video diffusion features 
with motion-aware subspaces. Most existing approaches either read 
out features at the clean endpoint of the host's generation or 
distill its weights into a separate model. PMD participates in the 
denoising process itself, decoding motion from intermediate states 
at every $\sigma$.

%% file: chapter/3method.tex
\section{Method}
\input{figure/main}

Motivated by our finding that large-scale video diffusion models 
implicitly carry a recoverable human motion signal along their 
denoising trajectories, we tackle text-to-motion generation by 
attaching a parasitic decoder to a frozen video model. Wan2.1, 
like other modern video diffusion models, operates as a latent 
diffusion model~\cite{rombach2022high}: a VAE~\cite{kingma2013auto} first encodes the video into a compact 
latent, and a DiT denoises this latent through a flow-matching 
schedule indexed by an interpolation coefficient 
$\sigma \in [0, 1]$. Given a text prompt $c$, the DiT advances the 
video latent from $\sigma=1$ (Gaussian noise) to $\sigma=0$ (clean 
video latent), which the VAE then decodes into a video 
$\mathbf{v}$. Our decoder runs in the same latent paradigm: it 
advances a motion latent in lockstep with Wan2.1 along the 
identical schedule, producing a 3D human motion $\mathbf{m}$ 
alongside the video.
We instantiate such a decoder in two stages 
(Figure~\ref{fig:method}). Stage 1 aligns a motion latent space 
with video latent space, giving the two streams a shared 
geometry. Stage 2 trains the parasitic motion decoder (PMD), a 
flow-matching decoder that shares Wan2.1's schedule and, at each 
step, reads features from selected DiT layers, fuses them through 
a $\sigma$-adaptive mechanism, and predicts a motion velocity.

\subsection{Stage 1: Motion-Video Latent Alignment}
\label{sec:stage1}

For the parasitic decoder to inherit the video model's generative 
trajectory, the motion representation must live in a space 
compatible with the video latent. Yet the two are architecturally 
and dimensionally far apart. Let 
$\mathbf{m} \in \mathbb{R}^{T \times d_m}$ denote a motion sequence 
in its raw kinematic representation, and 
$\mathbf{x'}^v_0 \in \mathbb{R}^{T_v \times H \times W \times C}$ 
the clean video latent produced by the video model's VAE, a 
high-dimensional spatial-temporal tensor designed for video 
reconstruction. Without first bridging this gap, the parasitic 
decoder would have to absorb the modality mismatch internally 
during flow matching, making training ill-conditioned. Stage 1 
closes the gap by learning a motion latent that aligns with a 
low-dimensional projection of the video latent 
(Figure~\ref{fig:method}a).

Stage 1 builds two parallel pipelines that meet in a shared latent 
space $\mathbb{R}^{T \times d_z}$. On the motion side, an encoder 
$E_m$ compresses $\mathbf{m}$ into 
$\mathbf{x}^m_0 = E_m(\mathbf{m}) \in \mathbb{R}^{T \times d_z}$, 
and a decoder $D_m$ reconstructs $\mathbf{m}$ from 
$\mathbf{x}^m_0$. Both are implemented as transformer stacks. On 
the video side, the frozen VAE first encodes the raw video into 
$\mathbf{x'}^v_0$, after which a projection head $P_v$ reduces it 
into the same space, 
$\mathbf{x}^v_0 = P_v(\mathbf{x'}^v_0) \in \mathbb{R}^{T \times d_z}$, 
using a per-frame spatial CNN to compress the spatial-channel 
dimensions into a $d_z$-dimensional feature, followed by temporal 
interpolation that resamples the sequence length to $T$.

The training objective combines three terms:
\begin{equation}
\mathcal{L}_{\text{stage 1}} 
= \underbrace{\|D_m(\mathbf{x}^m_0) - \mathbf{m}\|^2}_{\mathcal{L}_{\text{recon}}} 
+ \lambda_1 \underbrace{\|\mathbf{x}^m_0 - \mathbf{x}^v_0\|^2}_{\mathcal{L}_{\text{align}}}
+ \lambda_2 \underbrace{\|\mathbf{x}^m_0\|^2}_{\text{reg}},
\end{equation}
where $\mathcal{L}_{\text{recon}}$ preserves motion fidelity 
through the autoencoder, anchoring $\mathbf{x}^m_0$ to retain full 
motion information and preventing it from collapsing to a trivial 
solution. $\mathcal{L}_{\text{align}}$ pulls $\mathbf{x}^m_0$ 
toward $\mathbf{x}^v_0$, forcing $P_v$ to extract precisely those 
features of $\mathbf{x'}^v_0$ that carry motion information. The 
third term regularizes the latent magnitude.

\subsection{Stage 2: Parasitic Motion Decoder}
\label{sec:stage2}

With motion and video sharing the latent space 
$\mathbb{R}^{T \times d_z}$ from Stage 1, we now build the 
parasitic motion decoder (PMD) that denoises a motion latent in 
lockstep with the video DiT (Figure~\ref{fig:method}b). The 
central design choice in PMD is to make motion generation share 
the very same flow-matching schedule as video generation~\cite{lipman2023flow, liu2023flow, albergo2023building}, not 
merely a similar one. Tying the two processes to a single schedule 
lets motion and video align at every $\sigma$, so that features 
extracted from the host's denoising process can be consumed by 
motion denoising at the matched $\sigma$.

Concretely, the host video DiT~\cite{peebles2023scalable} learns a velocity field over its 
own video latent $\mathbf{x'}^v$ under a flow-matching schedule 
indexed by $\sigma \in [0, 1]$:
\begin{equation}
\resizebox{\linewidth}{!}{$\displaystyle
\mathbf{x'}^v_\sigma = (1-\sigma)\,\mathbf{x'}^v_0 + \sigma \boldsymbol{\epsilon}^v,
\quad
\mathbf{v}^{v*}_\sigma = \boldsymbol{\epsilon}^v - \mathbf{x'}^v_0,
\quad
\boldsymbol{\epsilon}^v \sim \mathcal{N}(0, \mathbf{I}).
$}
\end{equation}
We define motion denoising as a parallel process governed by the 
\emph{same} schedule on the aligned latent $\mathbf{x}^m_0$:
\begin{equation}
\resizebox{\linewidth}{!}{$\displaystyle
\mathbf{x}^m_\sigma = (1-\sigma)\,\mathbf{x}^m_0
  + \sigma \boldsymbol{\epsilon}^m,
\quad
\mathbf{v}^{m*}_\sigma = \boldsymbol{\epsilon}^m - \mathbf{x}^m_0,
\quad
\boldsymbol{\epsilon}^m \sim \mathcal{N}(0, \mathbf{I}).
$}\end{equation}
This shared formulation means that, for any $\sigma$, the host's 
intermediate state $\mathbf{x'}^v_\sigma$ and the motion state 
$\mathbf{x}^m_\sigma$ occupy the same index of their respective 
trajectories. The two noise draws $\boldsymbol{\epsilon}^v$ and 
$\boldsymbol{\epsilon}^m$ remain independent, so the coupling 
comes from the shared schedule and the feature conditioning rather 
than from a shared noise realization. PMD predicts 
$\mathbf{v}^{m*}_\sigma$ at every $\sigma$, conditioned on the 
text prompt $c$ and intermediate features hooked from the host 
DiT at the matched $\sigma$. Motion is thus produced not by an independent flow but by a decoder that shares the host's schedule and reads its features on the aligned latent space.

PMD is a lightweight flow-matching decoder built around a single 
guiding principle: rather than learn a generative trajectory from 
scratch, it follows the host's schedule and decodes the motion signal the host exposes at each step into an explicit motion. To 
do so, PMD must read this signal from \emph{within} the host's 
process. We therefore hook a set of intermediate features 
$\{\mathbf{f}_\ell\}_{\ell=1}^{L}$ from $L$ selected DiT layers at 
the matched $\sigma$, exposing the partially denoised 
representations on which the host's velocity prediction is built. 
These features are then fused into a single $\sigma$-dependent 
video signal $\mathbf{kv}_{\text{vid}}$ that PMD can directly 
attend to.

\subsubsection{$\sigma$-Adaptive Multi-Layer Fusion}
\label{para:fusion}
The hooked features $\{\mathbf{f}_\ell\}_{\ell=1}^{L}$ live in 
intermediate states of the video DiT and are not directly 
comparable either to one another or to the motion latent. To 
participate in motion denoising, they must first be brought into 
the motion latent space. We bring them into 
$\mathbb{R}^{T \times d_z}$ in three steps. First, each 
$\mathbf{f}_\ell$ is passed through the host DiT's own output 
head, mapping it from the intermediate-layer space to the final 
video latent space the head was trained for. Second, a small 
per-layer adapter corrects the residual mismatch of each layer. 
The host's head expects the final-layer distribution, and 
intermediate layers deviate from it. Finally, the projection 
$P_v$ from Stage 1 takes the adapted features into the shared 
motion-video latent space. $P_v$ is reused as is, with no further 
training. The first and third modules are therefore inherited 
from the host and Stage 1 respectively, and only the per-layer 
adapters are added in this stage.

The result is $L$ aligned video signals 
$\{\hat{\mathbf{f}}_\ell\}$ in $\mathbb{R}^{T \times d_z}$, ready 
to be combined and consumed by PMD. Layers at different depths 
expose the host's motion signal at different granularities, and 
their relative usefulness varies along the trajectory. A static 
fusion would commit the decoder to a single layer mixture across 
all $\sigma$. We instead let the fusion weights depend on 
$\sigma$:
\begin{equation}
\begin{aligned}
\mathbf{w}(\sigma)
  &= \mathrm{softmax}\bigl(\mathrm{MLP}(\mathbf{t}_\sigma)\bigr)
  \in \mathbb{R}^{L}, \\
\mathbf{kv}_{\text{vid}}
  &= \sum_{\ell=1}^{L} w_\ell(\sigma)\,\hat{\mathbf{f}}_\ell.
\end{aligned}
\end{equation}
The weights $\mathbf{w}(\sigma)$ are produced by a small MLP from 
the timestep feature $\mathbf{t}_\sigma$ and normalized over the 
layer index. The fused $\mathbf{kv}_{\text{vid}}$ is computed once 
per denoising step and shared across all decoder blocks. Trained 
end-to-end with the rest of PMD, $\mathbf{w}(\sigma)$ allows the 
decoder to discover, on its own, which layers to attend to at 
each $\sigma$.

\paragraph{Decoder blocks.}
With $\mathbf{kv}_{\text{vid}}$ in hand and text features 
$\mathbf{kv}_{\text{txt}}$ obtained from a frozen pretrained text 
encoder applied to $c$, PMD predicts the motion velocity 
$\hat{\mathbf{v}}^{m}_\sigma$ from the noisy motion latent 
$\mathbf{x}^m_\sigma$ at every $\sigma$. The decoder is realized 
as a stack of $N$ identical blocks operating on a hidden 
representation $\mathbf{h}$ obtained by linearly projecting 
$\mathbf{x}^m_\sigma$. Each block applies four sub-modules in 
sequence. Self-attention over the motion stream propagates 
information temporally within the motion trajectory. Video 
cross-attention to $\mathbf{kv}_{\text{vid}}$ transfers the host's 
denoising direction into the motion stream. Text cross-attention 
to $\mathbf{kv}_{\text{txt}}$ injects the semantics of the text 
prompt. A position-wise feed-forward network follows.

Every sub-module $f$ in PMD, together with the per-layer adapters 
and the fusion MLP described above, is conditioned on $\sigma$ 
through adaptive layer normalization (AdaLN):
\begin{equation}
\begin{aligned}
\bigl[\boldsymbol{\gamma},\,\boldsymbol{\beta},\,\boldsymbol{\alpha}\bigr]
  &= \mathrm{MLP}(\mathbf{t}_\sigma), \\
\mathbf{h}
  &\leftarrow \mathbf{h} + \boldsymbol{\alpha}\odot
  f\bigl(\boldsymbol{\gamma}\odot\mathrm{LN}(\mathbf{h})
  + \boldsymbol{\beta}\bigr).
\end{aligned}
\end{equation}
where $\boldsymbol{\gamma}$, $\boldsymbol{\beta}$ scale and shift 
the normalized input to $f$, and $\boldsymbol{\alpha}$ gates its 
residual contribution. This uniform $\sigma$-conditioning lets 
every operation in PMD adapt continuously to the current point on 
the schedule.

\paragraph{Training and inference.}
PMD is trained step-wise under the host's flow-matching schedule. 
At each iteration, we sample a clean motion latent $\mathbf{x}^m_0$ 
and its paired video latent $\mathbf{x'}^v_0$ from data, draw a 
coefficient $\sigma \sim \mathcal{U}[0, 1]$, and form 
$\mathbf{x}^m_\sigma$ and $\mathbf{x'}^v_\sigma$ via Eq.~(2) and 
Eq.~(3). The host DiT runs a forward pass on $\mathbf{x'}^v_\sigma$, 
from which the $L$ intermediate features $\{\mathbf{f}_\ell\}$ are 
hooked, and PMD predicts the motion velocity from 
$\mathbf{x}^m_\sigma$, $\sigma$, $c$, and $\{\mathbf{f}_\ell\}$, 
optimizing the standard flow-matching loss:
\begin{equation}
\mathcal{L}_{\text{stage 2}} 
= \mathbb{E}_{\sigma,\,\mathbf{x}^m_0,\,\boldsymbol{\epsilon}^m}
\bigl\|
\mathrm{PMD}\bigl(\mathbf{x}^m_\sigma, \sigma, c, \{\mathbf{f}_\ell\}\bigr) 
- \mathbf{v}^{m*}_\sigma
\bigr\|^2.
\end{equation}
The host DiT, the video VAE, the text encoder, and all Stage 1 
modules ($E_m$, $D_m$, $P_v$) are frozen. Only PMD's parameters, 
together with the per-layer adapters and the fusion MLP, are 
updated. At inference, no video latent is available in advance. 
PMD and the host DiT advance jointly from $\sigma=1$ to $\sigma=0$ 
along the shared schedule, with PMD reading the host's features at 
each step, after which $D_m$ decodes $\mathbf{x}^m_0$ into the 
final motion $\mathbf{m}$. A single joint pass thus yields the 
video from the host and the aligned motion from PMD.

%% file: figure/main.tex
\begin{figure*}[htbp]
  \centering
  \includegraphics[width=0.8\linewidth]{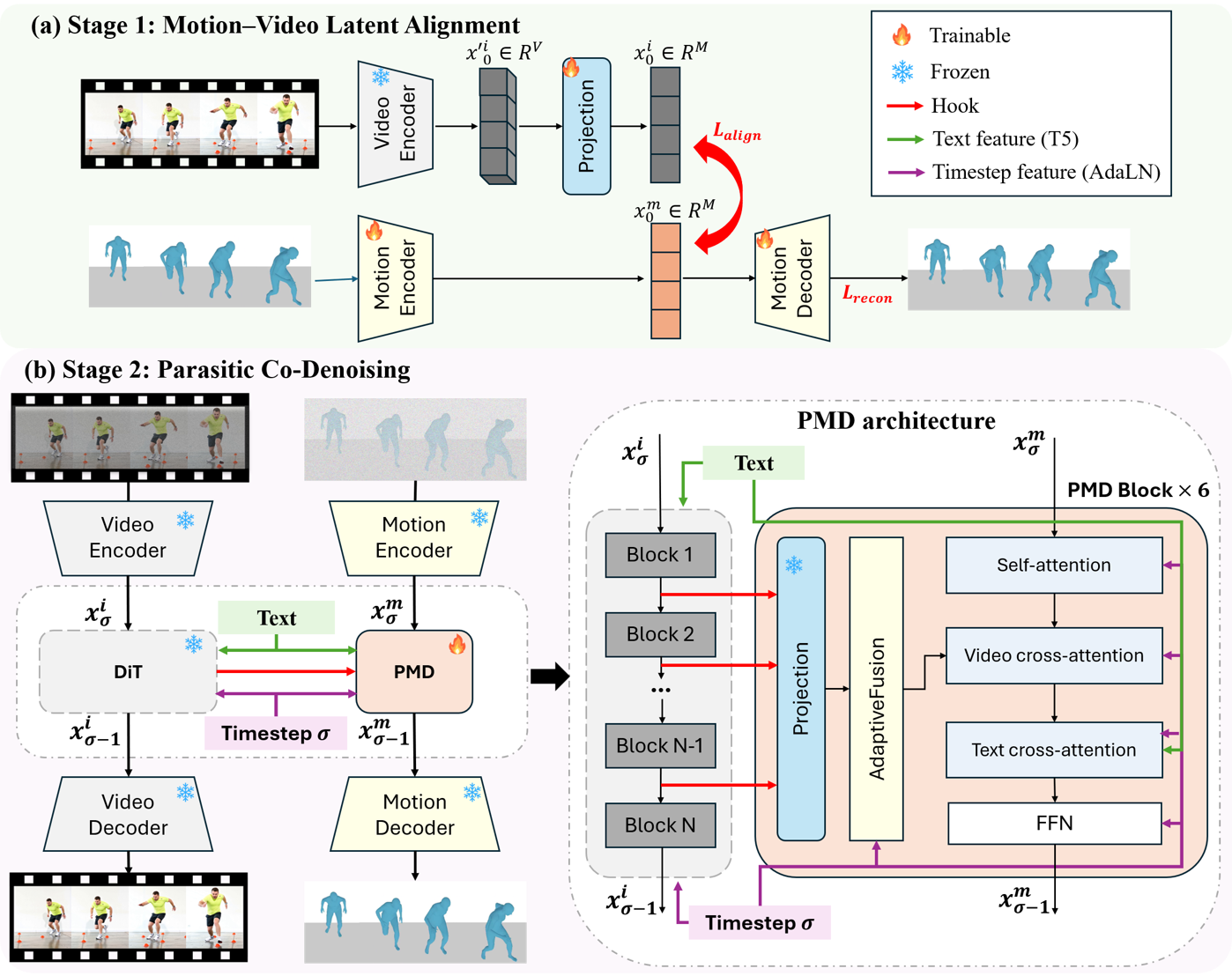}
  \vspace{-2mm}
  \caption{%
    The structure of Parasitic Motion Decoder (PMD).
  }
  \vspace{-5mm}
  \label{fig:method}
\end{figure*}

%% file: chapter/4experiment.tex
\section{Experiments}

\input{table/main_table}
\input{table/humanml}
\input{figure/visual}
\subsection{Setup}

\paragraph{Probing setup.}
We test whether a frozen video diffusion model carries motion 
along its denoising trajectory through a cross-modal alignment 
study. Each HumanML3D~\cite{guo2022generating} motion is rendered in 
Unreal Engine 5~\cite{karis2013real} from $10$ viewpoints with randomized identity, 
clothing, and background, leaving paired videos of the same motion with
 no shared appearance cues. At each $\sigma$, we extract 
velocity-field features from the host on a paired video and from 
HY-Motion~\cite{hymotion2025}, a SOTA text-to-motion model, on the 
corresponding motion. HY-Motion takes only motion as input, so 
alignment with its states reflects motion content, not visual 
similarity. We measure it by top-$1$ retrieval against $400$ 
held-out candidates, averaged over the ten viewpoints. 
Figure~\ref{fig:probing} shows a recoverable motion signal across 
the entire schedule, supporting parasitic co-denoising. We next 
turn this signal into competitive generation.

\input{table/ablation}

\paragraph{Generation setup.}
We evaluate on ViMoGen-228K~\cite{lin2025quest}: $171.5$K MoCap 
text-motion pairs, $56.6$K in-the-wild text-video-motion 
triplets, and $\sim$$11$K synthetic triplets. \textbf{Protocol A} 
trains on the synthetic subset ($10$K/$1$K train/eval), whose 
video distribution is closest to our host, isolating parasitic 
decoding from distribution shift. \textbf{Protocol B} trains on 
the full dataset and evaluates on MBench~\cite{lin2025quest}, the 
official benchmark, against the ViMoGen baselines. MBench scores 
two categories: semantic metrics (Motion-Condition Consistency, 
Motion Generalizability), rated by a vision-language model 
(Gemini Flash-2.5) on rendered videos, and kinematic metrics 
(Jitter Degree, Foot Sliding, Foot Floating, Dynamic Degree), 
computed on 3D joints.

\paragraph{Implementation details.}
The host is a frozen Wan2.1-T2V-14B~\cite{wan2025wan}. PMD has $N=6$ 
blocks, hooks $L=5$ DiT layers, hidden dimension $256$, and 
$\sim$$19$M trainable parameters. We train Stage 2 for $100$ 
epochs (Adam, lr $2 \times 10^{-4}$, batch $48$) with $0.1$ 
classifier-free guidance dropout applied independently to video 
and text. Inference runs Wan2.1's rectified-flow sampler for $50$ 
steps. All runs use $3$ NVIDIA RTX $4090$ GPUs.

\begin{figure}[t]
\centering
\includegraphics[width=0.9\linewidth]{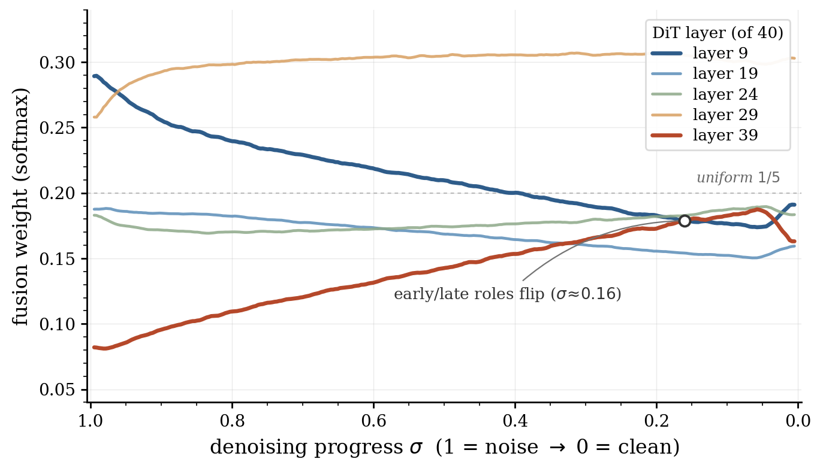}
\vspace{-2mm}
\caption{Learned $\sigma$-conditioned fusion weights
across $5$ hooked DiT layers (out of $40$). 
}
\vspace{-5mm}
\label{fig:fusion-weights}
\end{figure}

\subsection{Performance}
\label{sec:performance}

\subsubsection{Quantitative Results}
\label{sec:quantitative}

Table~\ref{tab:main} reports both protocols, and PMD leads the 
two semantic metrics in each. Under the data-scarce Protocol A, 
PMD outperforms every dedicated generator by a clear margin: when 
motion data is limited, reading the host's features along the schedule already supplies the semantic coverage that a from-scratch generator 
lacks. Under the large-scale Protocol B, PMD stays ahead of the 
far larger ViMoGen and the concurrent scene-conditioned 
MotionVLA, at a small fraction of their trainable parameters. The 
advantage thus holds from scarce to abundant data, tracking the 
host's representational depth rather than the size of any 
dedicated motion model.

\input{figure/alignment}

\paragraph{Qualitative results.}
Figure~\ref{fig:visual} compares PMD with dedicated generators on 
four MBench prompts. PMD follows the prompts more faithfully and 
produces larger, more expressive movement, most visibly on 
underrepresented actions such as \textit{Salsa dancing}, where 
the baselines fall back on generic motion. This long-tail 
coverage is inherited from the host through co-denoising. The 
bottom row shows the paired video the host renders alongside each 
motion, an output the baselines cannot produce at all.

\paragraph{Standard benchmark.}
Table~\ref{tab:humanml3d} evaluates PMD on HumanML3D, a benchmark 
of curated indoor MoCap. PMD leads on R-Precision, confirming its 
text-motion alignment on a second benchmark. Its FID trails the 
best MoCap-specialized methods, as expected: PMD inherits its 
distribution from the video host rather than fitting HumanML3D's 
narrow MoCap statistics, favoring semantic coverage over 
dataset-specific fit.

\subsection{Ablation}
\label{sec:ablation}

\paragraph{Paradigm-critical components.}
The clearest test of the paradigm is to remove the host signal: 
PMD collapses to a text-only generator of the same size 
(Table~\ref{tab:ablation}) and both semantic metrics fall sharply. 
The capability is the host's, not the decoder's. The other two 
components make it reachable. Removing the text condition keeps 
coarse semantics from the host but loses fine prompt control, and 
removing Stage~1 alignment is most damaging, leaving host features 
with no usable motion space and collapsing training.

\paragraph{Fusion strategy.}
We next examine how the host signal should be read. A single 
final-layer hook lags the full model by a wide margin, so the 
signal is not concentrated at the clean output but spread across 
the host's depth. A static uniform mixture narrows the gap yet 
still trails, since a fixed blend cannot track how layer relevance 
shifts along the schedule. Figure~\ref{fig:fusion-weights} exposes 
that shift: the mid-late layer L29 dominates throughout, while 
early- and late-layer contributions swap as denoising proceeds. 
Even hooking L29 alone, the strongest layer, falls short of the 
fusion, and it in turn beats a final-layer hook, so both which 
layer and how many matter. The motion signal lives in the shifting 
interplay across layers, and only $\sigma$-adaptive fusion tracks it.

\paragraph{Hooked layers.}
The same picture holds along depth. Shallow hooks underperform a 
spread across the host's full depth, and three full-depth layers 
already recover most of the gain while shallow ones cannot, so 
what matters is spanning the trajectory, not the layer count. 
Motion is present throughout the host's process, along both the 
$\sigma$ axis and the depth of its stack, and PMD works by reading 
it wherever it lives.
\input{figure/fusion_cluster}

\subsection{Discussion}

\paragraph{Co-denoising at inference.}
Figure~\ref{fig:flow-sync}(a) shows that the projected video 
latent $P_v(\mathbf{x}'^v_\sigma)$ and the motion latent 
$\mathbf{x}^m_\sigma$ grow progressively aligned along the 
schedule, peaking near $\sigma\!\approx\!0.5$ and staying high 
through $\sigma\!\to\!0$. Figure~\ref{fig:flow-sync}(b) visualizes this coupling: at every $\sigma$, the paired video and motion 
trajectories step in the same direction within a joint PCA basis. 
The motion side does not merely end up near the video, it moves 
with it throughout, a pattern consistent with co-denoising rather 
than post-hoc alignment.

\paragraph{Generative diversity under host guidance.}
Reading motion from the host invites a natural worry: that PMD 
merely transcribes each video into one fixed pose sequence, 
reducing generation to deterministic readout. 
Figure~\ref{fig:diversity} rules this out. For each host video, 
eight noise seeds yield distinct yet host-consistent motions, 
tightly clustered per video but clearly spread within each cluster 
(within-video pairwise MPJPE $40 \pm 12$~mm across $50$ videos). 
The host shapes the motion without fixing it, and PMD generates 
from its own noise rather than reading a predetermined answer off 
the host.

%% file: table/main_table.tex
\begin{table*}[t]
\centering

\small
\renewcommand{\arraystretch}{1.15}
\setlength{\tabcolsep}{4pt}
\resizebox{0.88\linewidth}{!}{%
\begin{tabular}{l c cc cccc}
\toprule
\multirow{2}{*}{\textbf{Method}} 
& \multirow{2}{*}{\textbf{\makecell{Trainable\\Params}}} 
& \multicolumn{2}{c}{\textbf{Semantic}} 
& \multicolumn{4}{c}{\textbf{Motion Quality}} \\
\cmidrule(lr){3-4} \cmidrule(lr){5-8}
& 
& \makecell{Motion\,Cond.\\Consistency\,$\uparrow$} 
& \makecell{Motion\\Generalizability\,$\uparrow$} 
& \makecell{Jitter\\Degree\,$\downarrow$} 
& \makecell{Dynamic\\Degree\,$\uparrow$} 
& \makecell{Foot\\Floating\,$\downarrow$} 
& \makecell{Foot\\Sliding\,$\downarrow$} \\
\midrule
\rowcolor{gray!8}
\multicolumn{8}{l}{\textit{Protocol A}} \\
\midrule
MDM~\cite{tevet2022human}                     & 23M    & 0.34 & 0.68 & \textbf{0.015} & 0.027 & 0.37 & 0.006 \\
T2M-GPT~\cite{zhang2023generating}              & 228M   & 0.34 & 0.74 & 0.035 & 0.034 & 0.33 & 0.005 \\
FineMoGen~\cite{zhang2023finemogen}         & 55M  & 0.29 & 0.63 & 0.031 & 0.033 & 0.39 & 0.007 \\
MotionLCM~\cite{dai2024motionlcm}         & 30M  & 0.41 & 0.86 & 0.025 & 0.038 & 0.27 & 0.009 \\
MoMask~\cite{guo2024momask}               & 45M    & 0.37 & 0.82 & 0.029 & 0.035 & \textbf{0.21} & 0.007 \\
MotionDiffuse~\cite{zhang2024motiondiffuse} & 31M    & 0.28 & 0.76 & 0.038 & 0.035 & 0.31 & 0.008 \\
MotionCraft~\cite{bian2025motioncraft}     & 150M & 0.46 & 0.89 & 0.019 & 0.037 & 0.34 & \textbf{0.005} \\
\rowcolor{blue!7}
\textbf{PMD (ours)}                & \textbf{19M} & \textbf{0.57} & \textbf{0.96} & 0.017 & \textbf{0.041} & 0.31 & 0.006 \\
\midrule
\rowcolor{gray!8}
\multicolumn{8}{l}{\textit{Protocol B}} \\
\midrule
MDM~\cite{tevet2022human}                     & 23M    & 0.42 & 0.51 & 0.014 & 0.038 & 0.16          & 0.014 \\
T2M-GPT~\cite{zhang2023generating}              & 228M   & 0.39 & 0.38 & 0.016 & 0.035 & 0.21          & 0.017 \\
FineMoGen~\cite{zhang2023finemogen}         & 55M  & 0.37 & 0.42 & 0.012 & 0.039 & 0.28          & 0.009 \\
MotionLCM~\cite{dai2024motionlcm}         & 30M  & 0.48 & 0.55 & 0.022 & 0.044 & 0.19 & 0.020 \\
MoMask~\cite{guo2024momask}               & 45M    & 0.38 & 0.44 & 0.015 & 0.040 & 0.18          & 0.015 \\
MotionDiffuse~\cite{zhang2024motiondiffuse} & 31M    & 0.44 & 0.42 & 0.011 & 0.029 & \textbf{0.13} & 0.006 \\
MotionCraft~\cite{bian2025motioncraft}     & 150M & 0.42 & 0.45 & 0.013 & 0.042 & 0.40         & 0.009 \\
ViMoGen-light~\cite{lin2025quest}       & 600M & 0.47 & 0.55 & 0.013 & 0.029 & 0.16          & 0.005 \\
ViMoGen~\cite{lin2025quest}             & 1.2B & 0.53 & 0.68 & \textbf{0.011} & 0.025 & 0.20 & 0.006 \\
MotionVLA~\cite{zhang2026motionvla} & 2B &  0.55 & 0.66 & 0.011 & 0.042 & 0.15 & 0.005 \\
\rowcolor{blue!7}
\textbf{PMD (ours)}                & \textbf{19M} & \textbf{0.58} & \textbf{0.72} & 0.014 & \textbf{0.044} & 0.15 & \textbf{0.005} \\
\bottomrule
\end{tabular}%
}
\vspace{-2mm}
\caption{Quantitative comparison on MBench under two protocols.}

\label{tab:main}
\end{table*}

%% file: table/humanml.tex
\begin{table*}[!thp]
\centering

\small
\setlength{\tabcolsep}{5pt}
\renewcommand{\arraystretch}{1.15}
\resizebox{0.88\textwidth}{!}{%
\begin{tabular}{l| ccc |cc c}
\toprule
Method & R-P Top-1$\uparrow$ & R-P Top-2$\uparrow$ & R-P Top-3$\uparrow$ 
& FID$\downarrow$ & MM-Dist$\downarrow$ & MModality$\uparrow$ \\
\midrule
MotionDiffuse~\cite{zhang2024motiondiffuse} & $0.491^{\pm.001}$ & $0.681^{\pm.001}$ & $0.782^{\pm.001}$ & $0.630^{\pm.001}$ & $3.113^{\pm.001}$ & $1.553^{\pm.042}$ \\
T2M-GPT~\cite{zhang2023generating}       & $0.492^{\pm.003}$ & $0.679^{\pm.002}$ & $0.775^{\pm.002}$ & $0.141^{\pm.004}$ & $3.121^{\pm.008}$ & $1.831^{\pm.048}$ \\
FineMoGen~\cite{zhang2023finemogen}     & $0.504^{\pm.002}$ & $0.690^{\pm.002}$ & $0.784^{\pm.002}$ & $0.151^{\pm.008}$ & $2.998^{\pm.008}$ & $\mathbf{2.696}^{\pm.079}$ \\
MoMask~\cite{guo2024momask}       & $0.521^{\pm.002}$ & $0.713^{\pm.002}$ & $0.807^{\pm.002}$ & $0.045^{\pm.002}$ & $2.958^{\pm.008}$ & $1.241^{\pm.040}$ \\
DisCoRD~\cite{cho2025discord}     & $0.524^{\pm.003}$ & $0.715^{\pm.003}$ & $0.809^{\pm.002}$ & $\mathbf{0.032}^{\pm.002}$ & $2.938^{\pm.010}$ & $1.288^{\pm.043}$ \\
GenM3~\cite{shi2025genm3}         & $0.511^{\pm.002}$ & $0.705^{\pm.002}$ & $0.804^{\pm.002}$ & $0.046^{\pm.002}$ & $\mathbf{2.852}^{\pm.009}$ & -- \\
MG-MotionLLM~\cite{wu2025mg}  & $0.516^{\pm.002}$ & $0.706^{\pm.002}$ & $0.802^{\pm.002}$ & $0.303^{\pm.010}$ & $2.952^{\pm.008}$ & $2.125^{\pm.159}$ \\
\midrule
\rowcolor{blue!7}
\textbf{PMD (ours)} & $\mathbf{0.537}^{\pm.002}$ & $\mathbf{0.733}^{\pm.002}$ & $\mathbf{0.831}^{\pm.002}$ & $0.098^{\pm.011}$ & $2.865^{\pm.007}$ & $1.987^{\pm.089}$ \\
\bottomrule
\end{tabular}%
}
\vspace{-2mm}
\caption{Text-to-motion results on HumanML3D.}

\label{tab:humanml3d}
\end{table*}

%% file: figure/visual.tex
\begin{figure*}[htbp]
  \centering
  
  \includegraphics[width=\linewidth]{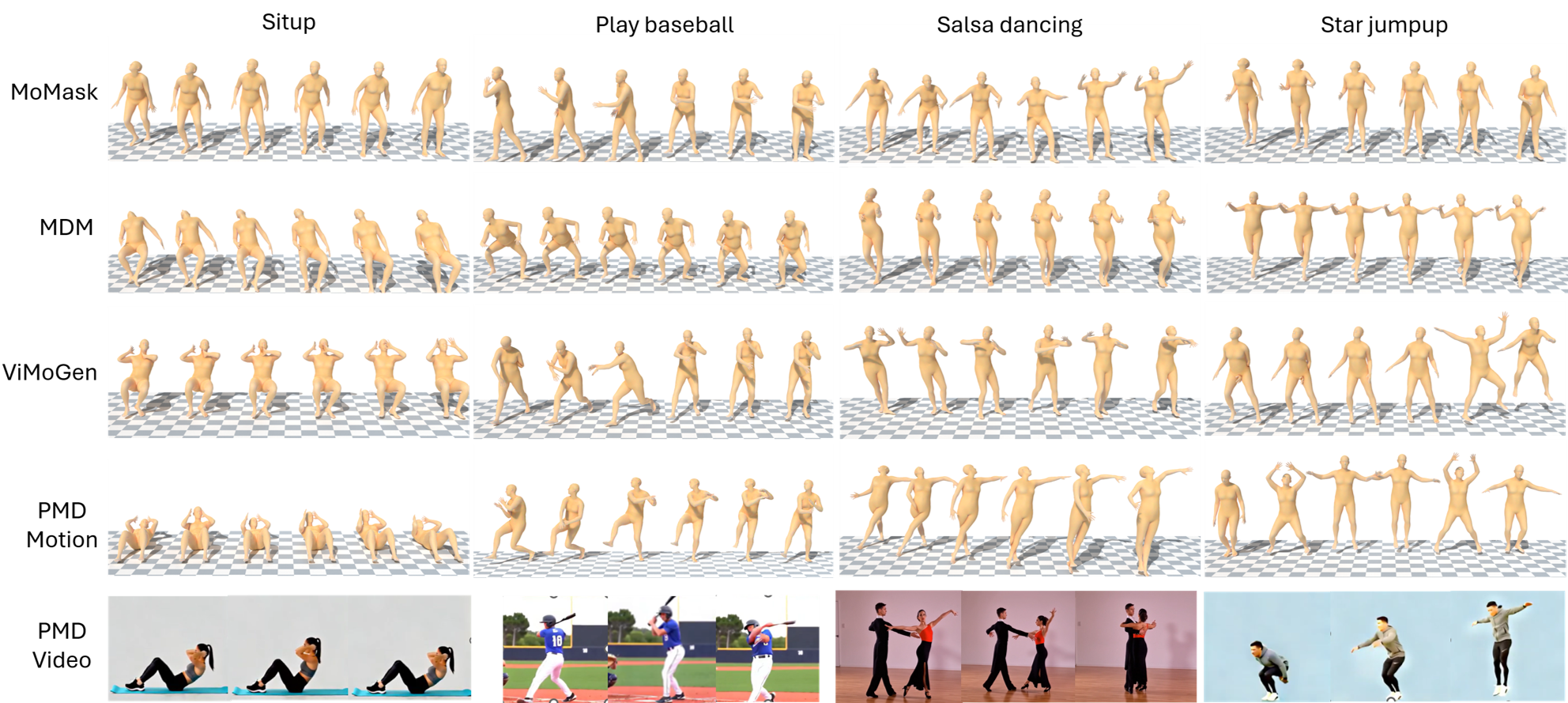}
  \vspace{-3mm}
  \caption{%
    Qualitative comparison on MBench prompts.
  }
  \vspace{-5mm}
  \label{fig:visual}
\end{figure*}

%% file: table/ablation.tex


\begin{table}[t]
\centering

\small
\setlength{\tabcolsep}{4pt}
\renewcommand{\arraystretch}{1.2}
\resizebox{0.95\columnwidth}{!}{%
\begin{tabular}{l cccc}
\toprule
Variant & MCC$\uparrow$ & MG$\uparrow$ & Jitter$\downarrow$ & FS$\downarrow$ \\
\midrule
\rowcolor{gray!15}
\multicolumn{5}{l}{\textit{Group 1: Paradigm-critical components}} \\
w/o host signal            & 0.29 & 0.46 & 0.037 & 0.011 \\
w/o text condition         & 0.42 & 0.74 & 0.022 & 0.007 \\
w/o Stage 1 align          & 0.21 & 0.38 & 0.045 & 0.015 \\
\midrule
\rowcolor{gray!15}
\multicolumn{5}{l}{\textit{Group 2: Fusion strategy}} \\
w/o fusion (final layer)   & 0.43 & 0.69 & 0.026 & 0.007 \\
Static (uniform)           & 0.47 & 0.75 & 0.021 & 0.008 \\
Single best layer (L29)    & 0.46 & 0.72 & 0.026 & 0.008 \\
\midrule
\rowcolor{gray!15}
\multicolumn{5}{l}{\textit{Group 3: Hooked layers}} \\
Shallow $\{1,3,5,7,9\}$    & 0.50 & 0.87 & 0.022 & 0.006 \\
$3$ layers $\{9,24,39\}$   & 0.52 & 0.89 & 0.024 & 0.008 \\
$7$ layers $\{9,15,19,24,29,34,39\}$ & 0.56 & 0.96 & 0.016 & 0.005 \\
\midrule
\rowcolor{blue!7}
\textbf{Full PMD}          & \textbf{0.57} & \textbf{0.96} & 0.017 & 0.006 \\
\bottomrule
\end{tabular}%
}
\vspace{-2mm}
\caption{Ablation on Protocol A. Groups vary paradigm-critical components~1, fusion strategy~2, and hooked layers~3. 
}

\label{tab:ablation}
\end{table}

%% file: figure/alignment.tex
\begin{figure*}[htbp]
  \centering
  \includegraphics[width=0.9\linewidth]{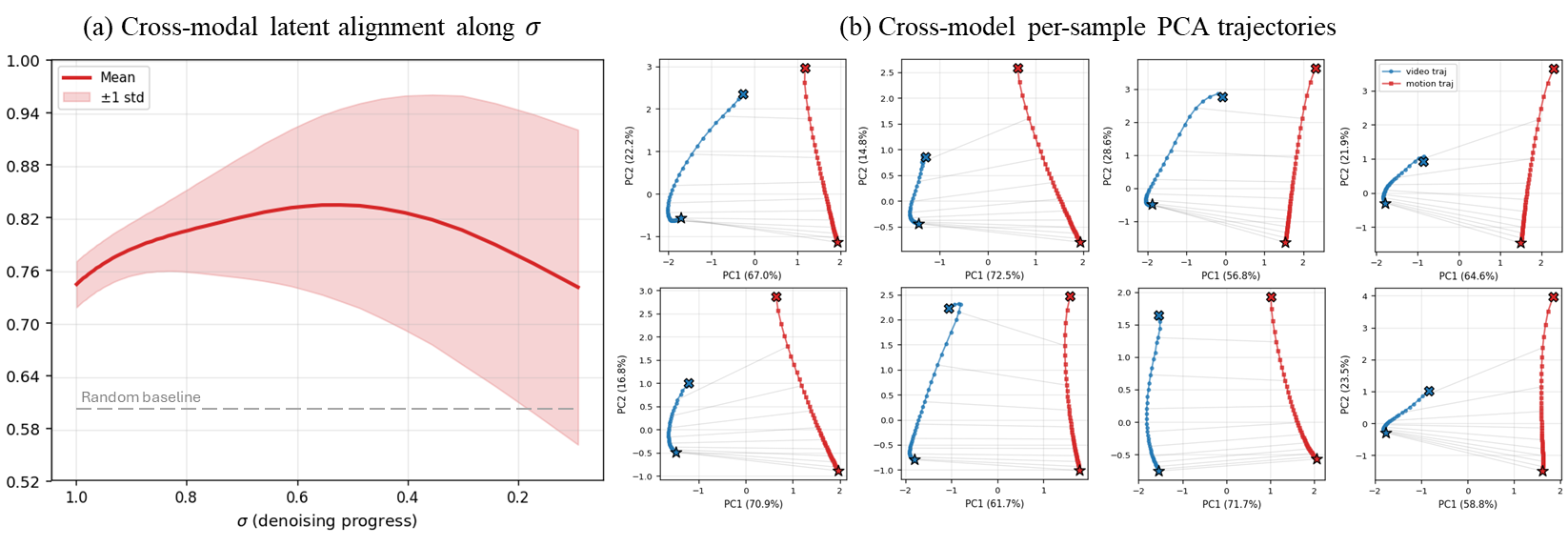}
  \vspace{-3mm}
 \caption{Video and motion latents along the denoising 
trajectory.
(a) Centered cosine similarity between 
$P_v(\mathbf{x'}^v_\sigma)$ and $\mathbf{x}^m_\sigma$, averaged 
over $184$ evaluation prompts. Shaded band shows $\pm 1$ std.
(b) Per-sample PCA projections of paired video (blue) 
and motion (red) latents from $\sigma=1$ ($\bigstar$) to 
$\sigma=0$ ($\times$).}
\vspace{-5mm}
\label{fig:flow-sync}
\end{figure*}

%% file: figure/fusion_cluster.tex
\begin{figure}[t]
\centering
\includegraphics[width=0.8\linewidth]{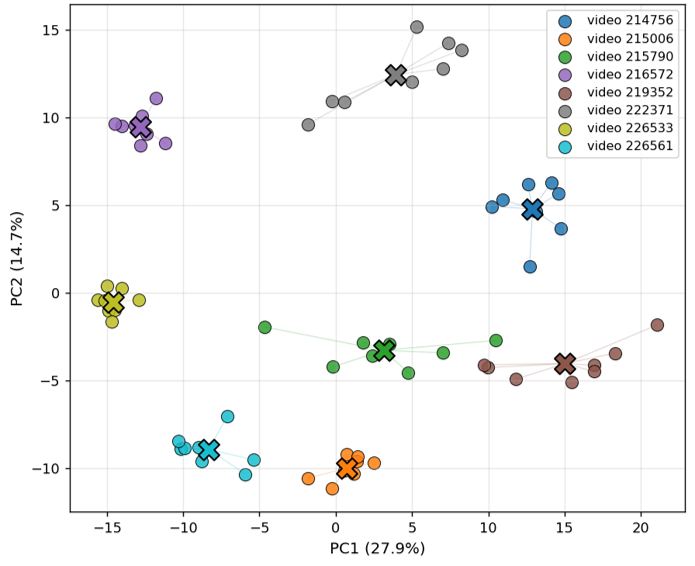}
\vspace{-3mm}
\caption{Per-video diversity of PMD-generated motions.
}
\vspace{-5mm}
\label{fig:diversity}
\end{figure}

%% file: chapter/5conclusion.tex
\section{Conclusion}

We presented parasitic co-denoising, a paradigm for generating 
$3$D human motion by decoding it from a frozen video diffusion model along its denoising schedule rather than 
training an independent motion generator. Our instantiation, the 
Parasitic Motion Decoder, attaches a flow-matching decoder 
to a frozen video host through $\sigma$-adaptive multi-layer 
fusion. At a small fraction of the trainable parameters of 
dedicated motion generators, it leads them on text-motion 
alignment. We hope this paradigm extends to other modalities whose 
generation is implicit in larger generative processes.

%% file: aaai2027.bib
@inproceedings{guo2022generating,
  title={Generating diverse and natural 3d human motions from text},
  author={Guo, Chuan and Zou, Shihao and Zuo, Xinxin and Wang, Sen and Ji, Wei and Li, Xingyu and Cheng, Li},
  booktitle={Proceedings of the IEEE/CVF conference on computer vision and pattern recognition},
  pages={5152--5161},
  year={2022}
}

@inproceedings{feng2024latent,
  title={Latent diffusion transformer for probabilistic time series forecasting},
  author={Feng, Shibo karis2013realnyan and Zhang, Zhong and Zhao, Peilin},
  booktitle={Proceedings of the AAAI Conference on Artificial Intelligence},
  volume={38},
  number={11},
  pages={11979--11987},
  year={2024}
}

@inproceedings{petrovich2022temos,
  title={{TEMOS}: Generating Diverse Human Motions from Textual Descriptions},
  author={Petrovich, Mathis and Black, Michael J. and Varol, G{\"u}l},
  booktitle={European Conference on Computer Vision (ECCV)},
  pages={480--497},
  year={2022},
  organization={Springer}
}

@inproceedings{lipman2023flow,
  title={Flow Matching for Generative Modeling},
  author={Lipman, Yaron and Chen, Ricky T. Q. and Ben-Hamu, Heli and Nickel, Maximilian and Le, Matthew},
  booktitle={The Eleventh International Conference on Learning Representations (ICLR)},
  year={2023}
}

@inproceedings{liu2023flow,
  title={Flow Straight and Fast: Learning to Generate and Transfer Data with Rectified Flow},
  author={Liu, Xingchao and Gong, Chengyue and Liu, Qiang},
  booktitle={The Eleventh International Conference on Learning Representations (ICLR)},
  year={2023}
}

@inproceedings{albergo2023building,
  title={Building Normalizing Flows with Stochastic Interpolants},
  author={Albergo, Michael S. and Vanden-Eijnden, Eric},
  booktitle={The Eleventh International Conference on Learning Representations (ICLR)},
  year={2023}
}

@inproceedings{peebles2023scalable,
  title={Scalable Diffusion Models with Transformers},
  author={Peebles, William and Xie, Saining},
  booktitle={Proceedings of the IEEE/CVF International Conference on Computer Vision (ICCV)},
  pages={4195--4205},
  year={2023}
}

@inproceedings{rombach2022high,
  title={High-Resolution Image Synthesis with Latent Diffusion Models},
  author={Rombach, Robin and Blattmann, Andreas and Lorenz, Dominik and Esser, Patrick and Ommer, Bj{\"o}rn},
  booktitle={Proceedings of the IEEE/CVF Conference on Computer Vision and Pattern Recognition (CVPR)},
  pages={10684--10695},
  year={2022}
}

@article{yang2024cogvideox,
  title={{CogVideoX}: Text-to-Video Diffusion Models with An Expert Transformer},
  author={Yang, Zhuoyi and Teng, Jiayan and Zheng, Wendi and Ding, Ming and Huang, Shiyu and Xu, Jiazheng and Yang, Yuanming and Hong, Wenyi and Zhang, Xiaohan and Feng, Guanyu and Yin, Da and Zhang, Xiaotao and Wang, Weihan and Cheng, Yean and Xu, Bin and Gu, Xiaotao and Dong, Yuxiao and Tang, Jie},
  journal={arXiv preprint arXiv:2408.06072},
  year={2024}
}

@article{kong2024hunyuanvideo,
  title={{HunyuanVideo}: A Systematic Framework for Large Video Generative Models},
  author={Kong, Weijie and Tian, Qi and Zhang, Zijian and Min, Rox and Dai, Zuozhuo and Zhou, Jin and Xiong, Jiangfeng and Li, Xin and Wu, Bo and Zhang, Jianwei and others},
  journal={arXiv preprint arXiv:2412.03603},
  year={2024}
}

@inproceedings{tang2023emergent,
  title={Emergent Correspondence from Image Diffusion},
  author={Tang, Luming and Jia, Menglin and Wang, Qianqian and Phoo, Cheng Perng and Hariharan, Bharath},
  booktitle={Advances in Neural Information Processing Systems (NeurIPS)},
  volume={36},
  year={2023}
}

@inproceedings{ke2024marigold,
  title={Repurposing Diffusion-Based Image Generators for Monocular Depth Estimation},
  author={Ke, Bingxin and Obukhov, Anton and Huang, Shengyu and Metzger, Nando and Daudt, Rodrigo Caye and Schindler, Konrad},
  booktitle={Proceedings of the IEEE/CVF Conference on Computer Vision and Pattern Recognition (CVPR)},
  year={2024}
}

@inproceedings{guo2022tm2t,
  title={{TM2T}: Stochastic and Tokenized Modeling for the Reciprocal Generation of 3D Human Motions and Texts},
  author={Guo, Chuan and Zuo, Xinxin and Wang, Sen and Cheng, Li},
  booktitle={European Conference on Computer Vision (ECCV)},
  pages={580--597},
  year={2022},
  organization={Springer}
}

@article{zhou2023metafi++,
  title={MetaFi++: WiFi-enabled transformer-based human pose estimation for metaverse avatar simulation},
  author={Zhou, Yunjiao and Huang, He and Yuan, Shenghai and Zou, Han and Xie, Lihua and Yang, Jianfei},
  journal={IEEE Internet of Things Journal},
  volume={10},
  number={16},
  pages={14128--14136},
  year={2023},
  publisher={IEEE}
}

@article{karis2013real,
  title={Real shading in unreal engine 4},
  author={Karis, Brian and Games, Epic},
  journal={Proc. Physically Based Shading Theory Practice},
  volume={4},
  number={3},
  pages={1},
  year={2013}
}

@article{kingma2013auto,
  title={Auto-encoding variational bayes},
  author={Kingma, Diederik P and Welling, Max},
  journal={arXiv preprint arXiv:1312.6114},
  year={2013}
}

@inproceedings{yuan2023physdiff,
  title={{PhysDiff}: Physics-Guided Human Motion Diffusion Model},
  author={Yuan, Ye and Song, Jiaming and Iqbal, Umar and Vahdat, Arash and Kautz, Jan},
  booktitle={Proceedings of the IEEE/CVF International Conference on Computer Vision (ICCV)},
  pages={16010--16021},
  year={2023}
}

@inproceedings{xie2024omnicontrol,
  title={{OmniControl}: Control Any Joint at Any Time for Human Motion Generation},
  author={Xie, Yiming and Jampani, Varun and Zhong, Lei and Sun, Deqing and Jiang, Huaizu},
  booktitle={The Twelfth International Conference on Learning Representations (ICLR)},
  year={2024}
}

@inproceedings{jiang2023motiongpt,
  title={{MotionGPT}: Human Motion as a Foreign Language},
  author={Jiang, Biao and Chen, Xin and Liu, Wen and Yu, Jingyi and Yu, Gang and Chen, Tao},
  booktitle={Advances in Neural Information Processing Systems (NeurIPS)},
  volume={36},
  year={2023}
}

@inproceedings{shi2025genm3,
  title={GenM3: Generative Pretrained Multi-path Motion Model for Text Conditional Human Motion Generation},
  author={Shi, Junyu and Liu, Lijiang and Sun, Yong and Zhang, Zhiyuan and Zhou, Jinni and Nie, Qiang},
  booktitle={Proceedings of the IEEE/CVF International Conference on Computer Vision},
  pages={13129--13139},
  year={2025}
}

@inproceedings{wu2025mg,
  title={Mg-motionllm: A unified framework for motion comprehension and generation across multiple granularities},
  author={Wu, Bizhu and Xie, Jinheng and Shen, Keming and Kong, Zhe and Ren, Jianfeng and Bai, Ruibin and Qu, Rong and Shen, Linlin},
  booktitle={Proceedings of the Computer Vision and Pattern Recognition Conference},
  pages={27849--27858},
  year={2025}
}

@inproceedings{cho2025discord,
  title={DisCoRD: Discrete Tokens to Continuous Motion via Rectified Flow Decoding},
  author={Cho, Jungbin and Kim, Junwan and Kim, Jisoo and Kim, Minseo and Kang, Mingu and Hong, Sungeun and Oh, Tae-Hyun and Yu, Youngjae},
  booktitle={Proceedings of the IEEE/CVF International Conference on Computer Vision},
  pages={14602--14612},
  year={2025}
}

@article{bhowmik2025moalign,
  title={Moalign: Motion-centric representation alignment for video diffusion models},
  author={Bhowmik, Aritra and Korzhenkov, Denis and Snoek, Cees GM and Habibian, Amirhossein and Ghafoorian, Mohsen},
  journal={arXiv preprint arXiv:2510.19022},
  year={2025}
}

@inproceedings{dai2024motionlcm,
  title={Motionlcm: Real-time controllable motion generation via latent consistency model},
  author={Dai, Wenxun and Chen, Ling-Hao and Wang, Jingbo and Liu, Jinpeng and Dai, Bo and Tang, Yansong},
  booktitle={European Conference on Computer Vision},
  pages={390--408},
  year={2024},
  organization={Springer}
}

@inproceedings{xiang2023denoising,
  title={Denoising diffusion autoencoders are unified self-supervised learners},
  author={Xiang, Weilai and Yang, Hongyu and Huang, Di and Wang, Yunhong},
  booktitle={Proceedings of the IEEE/CVF International Conference on Computer Vision},
  pages={15802--15812},
  year={2023}
}

@article{baranchuk2021label,
  title={Label-efficient semantic segmentation with diffusion models},
  author={Baranchuk, Dmitry and Rubachev, Ivan and Voynov, Andrey and Khrulkov, Valentin and Babenko, Artem},
  journal={arXiv preprint arXiv:2112.03126},
  year={2021}
}

@inproceedings{bian2025motioncraft,
  title={Motioncraft: Crafting whole-body motion with plug-and-play multimodal controls},
  author={Bian, Yuxuan and Zeng, Ailing and Ju, Xuan and Liu, Xian and Zhang, Zhaoyang and Liu, Wei and Xu, Qiang},
  booktitle={Proceedings of the AAAI Conference on Artificial Intelligence},
  volume={39},
  number={2},
  pages={1880--1888},
  year={2025}
}

@article{zhang2026motionvla,
  title={MotionVLA: Vision-Language-Action Model for Humanoid Motion},
  author={Zhang, Nonghai and Zhai, Siyu and Li, Yanjun and Zhang, Zeyu and Yin, Zhihan and Guo, Yandong and Shi, Boxin and Tang, Hao},
  journal={arXiv preprint arXiv:2606.15142},
  year={2026}
}

@article{lin2025quest,
  title={The quest for generalizable motion generation: Data, model, and evaluation},
  author={Lin, Jing and Wang, Ruisi and Lu, Junzhe and Huang, Ziqi and Song, Guorui and Zeng, Ailing and Liu, Xian and Wei, Chen and Yin, Wanqi and Sun, Qingping and others},
  journal={arXiv preprint arXiv:2510.26794},
  year={2025}
}

@article{zhang2023finemogen,
  title={Finemogen: Fine-grained spatio-temporal motion generation and editing},
  author={Zhang, Mingyuan and Li, Huirong and Cai, Zhongang and Ren, Jiawei and Yang, Lei and Liu, Ziwei},
  journal={Advances in Neural Information Processing Systems},
  volume={36},
  pages={13981--13992},
  year={2023}
}

@article{wan2025wan,
  title={Wan: Open and advanced large-scale video generative models},
  author={Wan, Team and Wang, Ang and Ai, Baole and Wen, Bin and Mao, Chaojie and Xie, Chen-Wei and Chen, Di and Yu, Feiwu and Zhao, Haiming and Yang, Jianxiao and others},
  journal={arXiv preprint arXiv:2503.20314},
  year={2025}
}

@article{hymotion2025,
  title={HY-Motion 1.0: Scaling Flow Matching Models for Text-To-Motion Generation},
  author={Tencent Hunyuan 3D Digital Human Team},
  journal={arXiv preprint arXiv:2512.23464},
  year={2025}
}

@article{yang2025echomotion,
  title={EchoMotion: Unified Human Video and Motion Generation via Dual-Modality Diffusion Transformer},
  author={Yang, Yuxiao and Sheng, Hualian and Cai, Sijia and Lin, Jing and Wang, Jiahao and Deng, Bing and Lu, Junzhe and Wang, Haoqian and Ye, Jieping},
  journal={arXiv preprint arXiv:2512.18814},
  year={2025}
}

@article{plappert2016kit,
  title={The kit motion-language dataset},
  author={Plappert, Matthias and Mandery, Christian and Asfour, Tamim},
  journal={Big data},
  volume={4},
  number={4},
  pages={236--252},
  year={2016},
  publisher={Mary Ann Liebert, Inc. 140 Huguenot Street, 3rd Floor New Rochelle, NY 10801 USA}
}

@article{blattmann2023stable,
  title={Stable video diffusion: Scaling latent video diffusion models to large datasets},
  author={Blattmann, Andreas and Dockhorn, Tim and Kulal, Sumith and Mendelevitch, Daniel and Kilian, Maciej and Lorenz, Dominik and Levi, Yam and English, Zion and Voleti, Vikram and Letts, Adam and others},
  journal={arXiv preprint arXiv:2311.15127},
  year={2023}
}

@article{duan2022survey,
  title={A survey of embodied ai: From simulators to research tasks},
  author={Duan, Jiafei and Yu, Samson and Tan, Hui Li and Zhu, Hongyuan and Tan, Cheston},
  journal={IEEE Transactions on Emerging Topics in Computational Intelligence},
  volume={6},
  number={2},
  pages={230--244},
  year={2022},
  publisher={IEEE}
}

@inproceedings{mourot2022survey,
  title={A survey on deep learning for skeleton-based human animation},
  author={Mourot, Lucas and Hoyet, Ludovic and Le Clerc, Fran{\c{c}}ois and Schnitzler, Fran{\c{c}}ois and Hellier, Pierre},
  booktitle={Computer Graphics Forum},
  volume={41},
  number={1},
  pages={122--157},
  year={2022},
  organization={Wiley Online Library}
}

@inproceedings{yu2026cmdm,
  title={Causal Motion Diffusion Models for Autoregressive Motion Generation},
  author={Yu, Qing and Watanabe, Atsushi and Fujiwara, Kent},
  booktitle={Proceedings of the IEEE/CVF Conference on Computer Vision and Pattern Recognition (CVPR)},
  year={2026}
}

@inproceedings{weng2026realign,
  title={ReAlign: Text-to-Motion Generation via Step-Aware Reward-Guided Alignment},
  author={Weng, Wanjiang and Tan, Xiaofeng and Wang, Junbo and Xie, Guo-Sen and Zhou, Pan and Wang, Hongsong},
  booktitle={Proceedings of the AAAI Conference on Artificial Intelligence (AAAI)},
  year={2026}
}

@inproceedings{zeng2025lightt2m,
  title={Light-T2M: A Lightweight and Fast Model for Text-to-motion Generation},
  author={Zeng, Ling-An and Huang, Guohong and Wu, Gaojie and Zheng, Wei-Shi},
  booktitle={Proceedings of the AAAI Conference on Artificial Intelligence (AAAI)},
  year={2025}
}

@inproceedings{guo2024momask,
  title={Momask: Generative masked modeling of 3d human motions},
  author={Guo, Chuan and Mu, Yuxuan and Javed, Muhammad Gohar and Wang, Sen and Cheng, Li},
  booktitle={Proceedings of the IEEE/CVF Conference on Computer Vision and Pattern Recognition},
  pages={1900--1910},
  year={2024}
}

@article{tevet2022human,
  title={Human motion diffusion model},
  author={Tevet, Guy and Raab, Sigal and Gordon, Brian and Shafir, Yonatan and Cohen-Or, Daniel and Bermano, Amit H},
  journal={arXiv preprint arXiv:2209.14916},
  year={2022}
}

@article{zhang2024motiondiffuse,
  title={Motiondiffuse: Text-driven human motion generation with diffusion model},
  author={Zhang, Mingyuan and Cai, Zhongang and Pan, Liang and Hong, Fangzhou and Guo, Xinying and Yang, Lei and Liu, Ziwei},
  journal={IEEE transactions on pattern analysis and machine intelligence},
  volume={46},
  number={6},
  pages={4115--4128},
  year={2024},
  publisher={IEEE}
}

@inproceedings{zhang2023generating,
  title={Generating human motion from textual descriptions with discrete representations},
  author={Zhang, Jianrong and Zhang, Yangsong and Cun, Xiaodong and Zhang, Yong and Zhao, Hongwei and Lu, Hongtao and Shen, Xi and Shan, Ying},
  booktitle={Proceedings of the IEEE/CVF conference on computer vision and pattern recognition},
  pages={14730--14740},
  year={2023}
}

@book{burdea2003virtual,
  title={Virtual reality technology},
  author={Burdea, Grigore C and Coiffet, Philippe},
  year={2003},
  publisher={John Wiley \& Sons}
}
